\documentclass[a4paper]{article}
\pdfoutput=1
\usepackage{INTERSPEECH2022}
\usepackage{multirow}
\usepackage{booktabs}
\usepackage{url}

\title{A Context-Aware Feature Fusion Framework for Punctuation Restoration}
\name{Yangjun Wu, Kebin Fang, Yao Zhao}
\address{
  Institute of Computing Innovation, Zhejiang University}
\email{\{yjwu, fkb, zyao\}@zjuici.com}

\begin{document}

\maketitle
\begin{abstract}
To accomplish the punctuation restoration task, most existing approaches focused on leveraging extra information (e.g., part-of-speech tags) or addressing the class imbalance problem. Recent works have widely applied the transformer-based language models and significantly improved their effectiveness. To the best of our knowledge, an inherent issue has remained neglected: the attention of individual heads in the transformer will be diluted or powerless while feeding the long non-punctuation utterances. Since those previous contexts, not the followings, are comparatively more valuable to the current position,  it's hard to achieve a good balance by independent attention. In this paper, we propose a novel \textbf{F}eature \textbf{F}usion framework based on two-type \textbf{A}ttentions (FFA) to alleviate the shortage. It introduces a two-stream architecture. One module involves interaction between attention heads to encourage the communication, and another masked attention module captures the dependent feature representation. Then, it aggregates two feature embeddings to fuse information and enhances context-awareness. The experiments on the popular benchmark dataset IWSLT demonstrate that our approach is effective. Without additional data, it obtains comparable performance to the current state-of-the-art models.
\end{abstract}


\section{Introduction}
Punctuation restoration is a significant post-processing step in automatic speech recognition (ASR) systems, because punctuation marks are not usually predicted. It can enhance the readability of speech transcripts and contribute to downstream tasks, such as machine translation, intent detection, or slot filling in dialogue systems.

The task attracts a large amount of interest, previous works can be generally categorized into three groups: 1) First, a portion of the studies \cite{7846300,klejch2017sequence,8682260,8545470} treat this problem as a machine translation task by feeding a non-punctuation sequence and yielding an output sequence with a mark. 2) Second, some researchers \cite{Adversarial, 9053159, Lin2020JointPO,alam-etal-2020-punctuation, Tilk2016BidirectionalRN,shi21_interspeech} regard it as a sequence labeling task, in which a punctuation mark is assigned to each word by the probability. 3) The others \cite{Che2016PunctuationPF} view it as a token-level classification task, forecasting a tag for each token via the classifier.

To address this problem, early studies\cite{gale17_interspeech, zelasko18_interspeech} typically used Long Short-Term Memory (LSTM) or Convolutional Neural Network (CNN) to capture the contextual information. More recently, transformer-based models are widely applied to remarkably boost the performance. Unfortunately, these methods only utilize the self-attention mechanism and put major effort into integrating external knowledge (e.g., part-of-speech tags), data augmentation, or handling the labels imbalance to tackle the punctuation predictions issue. Rarely emphasis has been placed on the limitation of self-attention itself. The experiments \cite{Bottleneck, Talking} show that the limitation called \textit{Low-Rank Bottleneck} undeniably exists in the self-attention mechanism. Concretely, with the fixed parameters size of self-attention layer\cite{attention}, increasing the number of heads would decrease the head size, which causes a decrease in the expressive power of each head. For instance, the dimension of Bert-base \cite{bert} in each head is 64, which is far less than the length of non-punctuation utterances (256). Meanwhile, the experiments \cite{Discriminative} indicate that the left context is relatively more vital than the right context to the current position. Some previous approaches used a sliding window with both left and right overlapped context to tackle this inherent obstacle. It could partially alleviate the shortcoming, but there have been no attempts to advance the transformer structure itself.

Inspired by these observations, we propose a \textbf{F}eature \textbf{F}usion framework based on two-type \textbf{A}ttentions (FFA) \footnote{The code, dataset, and evaluation results are public available at \url{https://github.com/Young1993/ffa}} to mitigate the barrier. Specifically, we design a two-stream structure, including a masked self-attentions-based module (MSA) to pay more attention to the previous tokens, another module (ISA) based on the interaction between self-attention heads to increase knowledge sharing. We first attain two-type feature representations via these two modules, respectively. We incorporate the two embeddings to aggregate feature information to achieve the context-aware at the fusing stage. Lastly, we yield token-level punctuation tags as the output. Our main contributions are summarized as follows:
\begin{itemize}
\item We present a novel framework, FFA, to encourage message sharing and capture the dependent feature to advance the shortage of the standard attention mechanism. 
\item Without extra data,  the results on the popular benchmark dataset IWSLT indicate that FFA can leverage the dataset itself and obtain comparable performance to the current state-of-the-art model, which demonstrates that FFA is effective.
\item We introduce a novel interaction self-attention-based module to share the information between heads. The ablation studies show that this module can increase the expression capability of attention heads and improve the robustness.
\end{itemize}

\section{Problem Definition}
Given the input sequence $ X = (x_1, x_2, ... x_n) $ and punctuation tags $Y = (y_1, y_2,... y_n)$, punctuation restoration is defined as a token-level classification task that outputs a sequence  $ \hat{Y} = (\hat{y}_1, \hat{y}_2, ... \hat{y}_n) $ in \textit{[O, COMMA, PERIOD, QUESTION]}, the \textit{O} denotes the label is None, where n is the length of the input utterance. During the testing stage, we automatically add the predicted mark to the position after the current token.

\section{Methodology}
\begin{figure}[t]
  \centering
  \includegraphics[width=\linewidth]{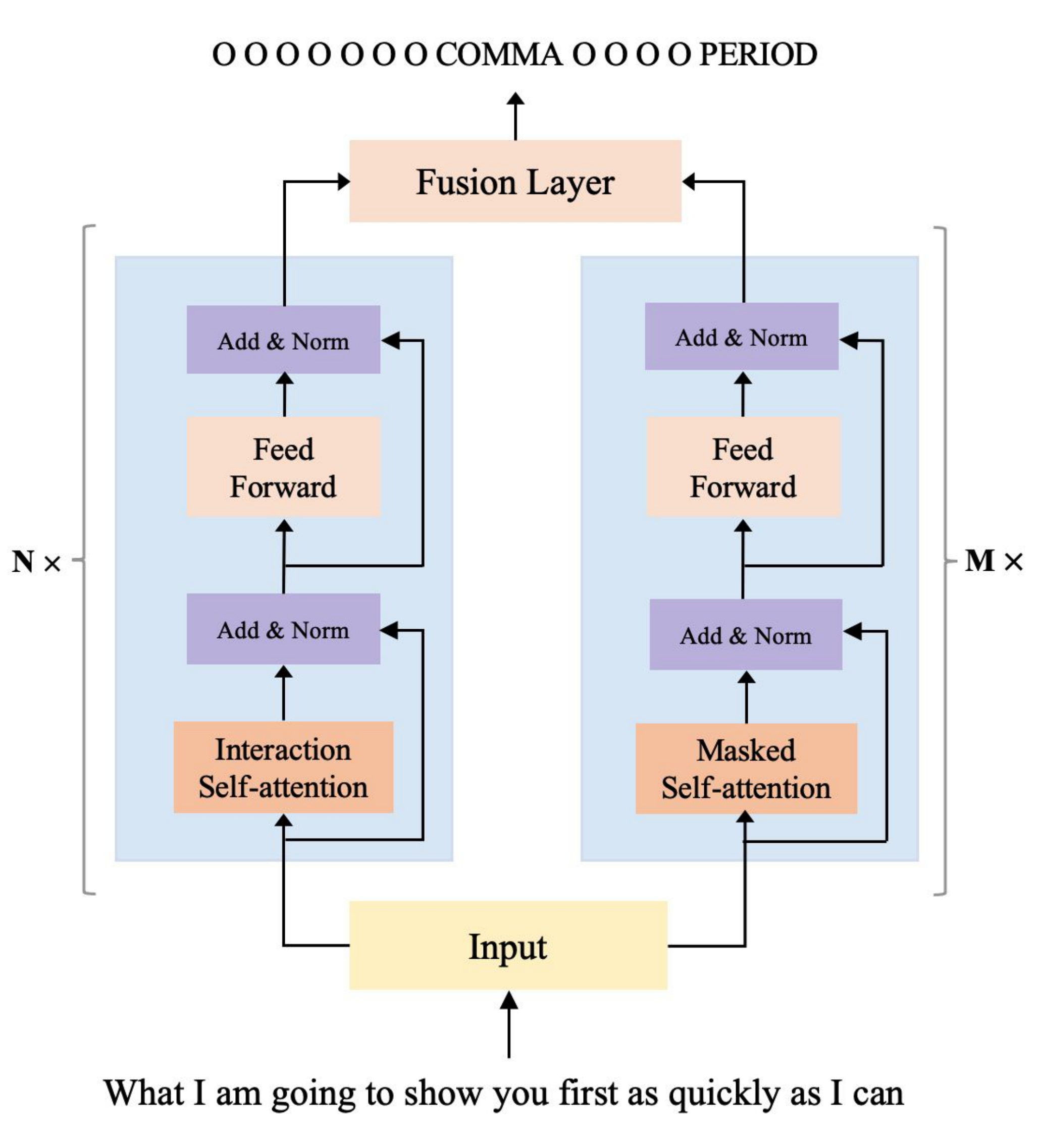}
  \caption{Overall architecture of FFA. N and M refer to N and M layers, respectively. }
  \label{fig:mdl architecture}
\end{figure}

In this section, we will formulate FFA in detail. As described in Figure \ref{fig:mdl architecture}, FFA contains three core components: Interaction Self-attention based module (ISA), Masked Self-attention based module (MSA), and Fusion Layer (FL).

\subsection{Initialization}
During the initialization phase, we first obtain the input embedding via a pre-trained language model (e.g., Funnel-Transformer\cite{funnel}, Bart\cite{bart}). Then, the initialized embedding is fed into ISA and MSA modules in parallel.

\subsection{Interaction Self-attention (ISA)}
\begin{figure}[t]
  \includegraphics[width=\linewidth]{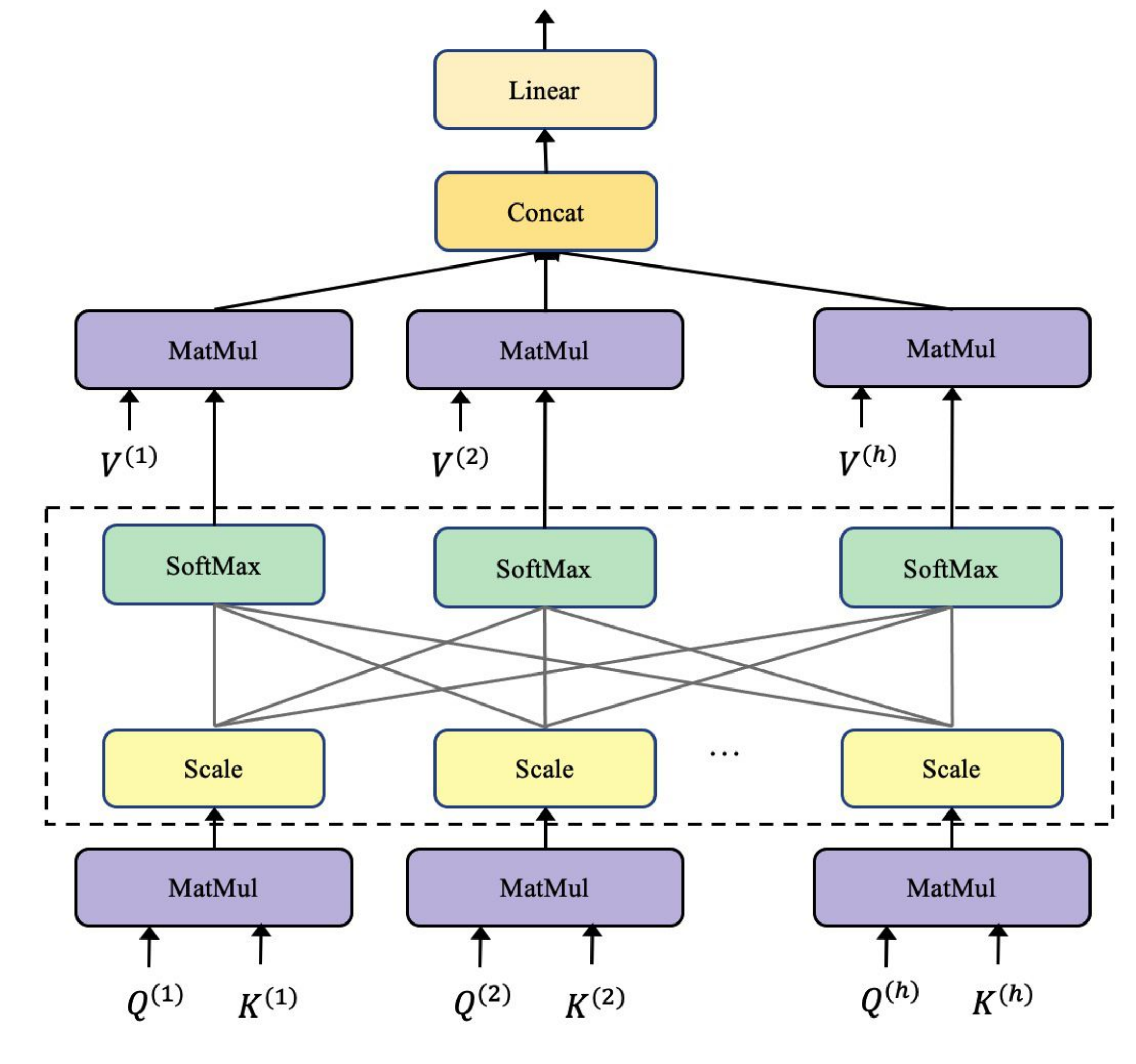}
  \caption{The calculation of interaction self-attention between Q, K and V in heads.}
  \label{fig:ISA architecture}
\end{figure}

This module aims to alleviate the inherent issue (\textit{Low-Rank Bottleneck}) in the Multi-Head self-attention mechanism. The Figure \ref{fig:ISA architecture} illustrates the detailed calculation of interaction self-attention.

To a specific head $k$, the conventional scaled dot-product attention\cite{attention} computation lies:
\begin{gather}
    Attn^{(k)}{(Q,K,V)} = softmax(\frac{J^{(k)}}{\sqrt{d_{emb} / H}})V^{(k)} \\
    J^{(k)} = {Q^{(k)}K^{(k)}}^\mathrm{T} \\
    Q = x_i W_Q^h, K = x_j W_K^h, V = x_j W_V^h\label{equa1}
\end{gather}

Here, $J^{(k)}$ denotes the dot-product between query $Q$ and key $K$ for the {$k$-th} head, $d_{emb}$ refers to the embedding size, $H$ is the head numbers. 

Unlike the vanilla self-attention, we design the interaction self-attention as a two-stage computation. It first attains the attention scores $J^{(k)}$ for individual heads. Then, it aggregates these scores to update the attention weights via a projection component $\phi$, which increases the feature interaction between heads to share the message.

For the component $\phi$, we introduce a new matrix $P_\lambda$ \footnote{$P_\lambda$ is initialized with standard deviations of $0.1/\sqrt{d_{emb}}$} to obtain extra information from sibling heads. Then, we update the $J^{(k)}$ as $\hat{J^{(k)}}$ to represent the new attention weights as below:

\begin{gather}
\begin{pmatrix}
\hat{J^{(1)}} \\\\
\hat{J^{(2)}} \\\\
\vdots \\\\
\hat{J^{(h)}} \\\\
\end{pmatrix} = P_\lambda\begin{pmatrix}
J^{(1)} \\\\
J^{(2)} \\\\
\vdots \\\\
J^{(h)} \\\\
\end{pmatrix} + 
\begin{pmatrix}
J^{(1)} \\\\
J^{(2)} \\\\
\vdots \\\\
J^{(h)} \\\\
\end{pmatrix},
P_\lambda = 
\begin{pmatrix}
    \lambda_{11} & \cdots & \lambda_{1h}\\\\
    \lambda_{21} & \cdots & \lambda_{2h}\\\\
    \vdots & \ddots &  \vdots \\\\
    \lambda_{h1} & \cdots &  \lambda_{hh}\\\\
\end{pmatrix}
\end{gather}

Here, $\lambda_{hh}$ denotes the learnable parameter. Now, we modify the Equation \ref{equa1} to novel Equation \ref{equa5} as the interaction scores:
\begin{equation}
    Attn^{(k)}{(Q,K,V)} = softmax(\frac{\hat{J^{(k)}}}{\sqrt{d_{emb} / H}})V^{(k)}\label{equa5}
\end{equation}
Finally, we repeat $h$ times the projections in different $Q, K, V $, concatenate the embeddings and employ a linear layer to yield the final matrices $M$ as follows:
\begin{equation}
 M\left(Q,K,V\right)= Concat\left({\rm Attn}_{(1)},\ldots,{\rm Attn}_{(h)}\right)W  + B
\end{equation}
Where $W$ is a learnable parameter and $B$ is the bias. After that, we follow the standard transformer workflow and feed the matrices into the layer normalization and feed-forward.

\begin{table*}[!ht]
  \caption{Results in terms of P(\%), R(\%), F1(\%) on the English IWSLT2011 test set. We collect the results from the original papers without modification, and the highest numbers are in bold.}
  \label{tab:1}
  \centering
  \setlength{\tabcolsep}{2.5mm}{
    \begin{tabular}{ c|r|r|r|r|r|r|r|r|r|r|r|r}
       \toprule
        \multirow{2}*{\textbf{Model}} & \multicolumn{3}{c|}{\textbf{Comma}} &\multicolumn{3}{c|}{\textbf{Period}} &\multicolumn{3}{c|}{\textbf{Question}} &\multicolumn{3}{c}{\textbf{Overall}} \\
        \cline{2-13} & P & R& F1 & P & R& F1 & P & R& F1 & P & R& F1\\
        \hline
        T-BRNN-pre & 65.5&47.1&54.8&73.3&72.5&72.9&70.7&63.0&66.7&70.0&59.7&64.4 \\ [1pt]
        BLSTM-CRF &58.9&59.1&59.0&68.9&72.1&70.5&71.8&60.6&65.7&66.5&63.9&65.1 \\ [1pt]
        Teacher-Ensemble&66.2&59.9&62.9&75.1 &73.7& 74.4&72.3&63.8&67.8&71.2&65.8&68.4\\ [1pt]
        DRNN-LWMA-pre& 62.9&60.8&61.9&77.3&73.7&75.5&69.6&69.6&69.6&69.9&67.2&68.6 \\ [1pt]
        Self-attention-word-speech& 67.4&61.1&64.1&82.5&77.4&79.9&80.1&70.2&74.8&76.7&69.6&72.9\\ [1pt]
        CT-Transformer &68.8&69.8&69.3&78.4&82.1&80.2&76.0&82.6&79.2&73.7&76.0&74.9\\ [1pt]
        SAPR & 57.2&50.8&55.9&\textbf{96.7}&\textbf{97.3}&\textbf{96.8}&70.6&69.2&70.3&78.2&74.4&77.4\\ [1pt]
        BERT-base+Adversarial&76.2&71.2&73.6&87.3&81.1&84.1&79.1&72.7&75.8&80.9&75.0&77.8\\[1pt]
        BERT-large+Transfer&70.8&74.3&72.5&84.9&83.3&84.1&82.7&93.5&87.8&79.5&83.7&81.4\\[1pt]
        BERT-base+FocalLoss&74.4&77.1&75.7&87.9&88.2&88.1&74.2&88.5&80.7&78.8&84.6&81.6\\[1pt]
        RoBERTa-large+augmentation&76.8&76.6&76.7&88.6&89.2&88.9&82.7&93.5&87.8&82.6&83.1&82.9\\[1pt]
        RoBERTa-base&76.9&75.4&76.2&86.1&89.3&87.7&88.9&87.0&87.9&84.0&83.9&83.9\\[1pt]
        RoBERTa-large+SCL&78.4&73.1&75.7&86.9&87.2&87.0&\textbf{89.1}&89.1&89.1&\textbf{84.8}&83.1&83.9\\[1pt]
        %Funnel-transformer-xlarge&75.5&82.4&78.8&88.7&89.0&88.9&82.4&91.3&86.6&81.7&85.8&83.7\\[1pt]
        FT+POS+SBS &78.9&78.0&78.4&86.5&93.4&89.8&87.5&91.3&\textbf{89.4}&82.9&85.7&84.3\\[1pt]
        ELECTRA-large+Disc-ST&78.0&\textbf{82.4}&80.1&89.9&90.8&90.4&79.6&93.5&86.0&83.6&\textbf{86.7}&\textbf{85.2}\\[1pt]
        \hline
        FFA - w/o ISA &75.8&76.6&76.2&86.7&89.8&88.2&81.6&87.0&84.2&81.3&83.2&82.2\\[1pt]
        FFA - w/o MSA &78.1&78.6&78.4&89.1&89.0&89.1&77.2&94.8&85.1&83.3&84.1&83.5\\[1pt]
        FFA           &\textbf{79.4}&81.1&\textbf{80.3}&89.8&90.7&90.2&77.2& \textbf{95.7} & 85.4 & 84.2 & 86.1 & \textbf{85.2} \\[1pt]
        \bottomrule
  \end{tabular}
  }
\end{table*}

\subsection{Masked Self-attention (MSA)}
To encourage the models to pay more attention to the left context other than the right, we follow the masked self-attention \cite{attention} involved in the transformer decoder, to avoid the information leakage for the current token. 

Unable to visualize the latter information, this module will focus on reasoning about the current token based on the previous elements, to preserve the auto-regressive property. Different to ISA, we implement this inside of the scaled dot-product self-attention via masking out (setting to $-\infty$) all values in the input of the softmax that correspond to illegal connections.

\subsection{Fusion Layer}
As we obtain the feature representation of $C_i$ and $C_m$ via ISA and MSA, respectively. Then,  $C_i$ and $C_m$ are concatenated as the fused representation $H\in\mathbb{R}^{n\times{(d_i+d}_{m)}}$, and the fused vector is fed into one layer of transformer to obtain the final output sequence $\hat{Y}$.

We use cross-entropy loss $\ell$ as follows:
\begin{equation}
    \ell = - \sum_{i=1}^{K} p_i\log p_i
\end{equation}
Here, K is the total number of categories, $p_i$ is the predicted probability of label $i$.

\section{Experiments}
In this section, we compare the performance of FFA with the state-of-the-art approaches on the popular dataset in the fair comparison setting and further ablate some design choices in FFA to understand their contributions.
\subsection{Dataset}
The English IWSLT2011\cite{IWSLT2012} is a popular benchmark dataset for punctuation restoration. It contains 2.1M words for training set, 296K words for validation set, and 12626 words for the manual transcription test set. It also contains 12822 words for the actual ASR transcription test set, whcih the words are predicted by ASR systems, so it would comprise some grammatical errors or wrong words. There are four types of punctuation marks (none, comma, period, and question mark), the distribution of categories in the training dataset is as follows: 85.7\% without punctuation, 7.53\% with comma, 6.3\% with period and 0.47\% with question marks. In this following, we will employ precision (P), recall (R), and F1-score (F1) to evaluate FFA and other approaches.

\subsection{Baselines}
We compare FFA with the top performer models. One part methods are RNN-based: T-BRNN-pre, BLSTM-CRF, Teacher-Ensemble, and DRNN-LWMA-pre (employs a deep recurrent neural network). The other part approaches are transformer-based, including Self-attention-word-speech, CT-Transformer, SAPR (uses a transformer encoder-decoder architecture and views the punctuation restoration as a translation task), BERT-base+Adversarial, BERT-large+Transfer, BERT-base+FocalLoss (employs the focal loss not cross entropy loss to advance the results), RoBERTa-large+augmentation, RoBERTa-base (predicts the tags by multiple context windows), RoBERTa-large+SCL (contrastive learning), funnel-transformer-xlarge+POS+Fusion+SBS abbreviated as FT+POS+SBS (incorporates an external POS tagger and fuses its predicted labels into the existing language model to address the problem), and ELECTRA-large+Disc-ST (introduces a discriminative self-training approach with weighted loss and uses external dataset to achieve the SOTA performance). \cite{Tilk2016BidirectionalRN,Yi2017DistillingKF,8682418,8682260,DBLP:journals/corr/abs-2003-01309,8545470,Adversarial,9023200,Yi2020,alam-etal-2020-punctuation,courtland-etal-2020-efficient,Shi2021IncorporatingEP,DBLP:journals/corr/abs-2107-09099, Discriminative}.
\begin{table*}[!ht]
  \caption{Results in terms of P(\%), R(\%), F1(\%) on the English IWSLT2011 ASR transcription test set.}
  \label{tab:2}
  \centering
  \setlength{\tabcolsep}{2.5mm}{
    \begin{tabular}{c|r|r|r|r|r|r|r|r|r|r|r|r}
       \toprule
        \multirow{2}*{\textbf{Model}} & \multicolumn{3}{c|}{\textbf{Comma}} &\multicolumn{3}{c|}{\textbf{Period}} &\multicolumn{3}{c|}{\textbf{Question}} &\multicolumn{3}{c}{\textbf{Overall}} \\
        \cline{2-13} & P & R& F1 & P & R& F1 & P & R& F1 & P & R& F1\\
        \hline
        Self-attention-word-speech&64.0&59.6&61.7&75.5&75.5&75.6&\textbf{72.6}&65.9&\textbf{69.1}&\textbf{70.7}&67.1& - \\[3pt]
        BERT-base+Adversarial&\textbf{70.7}&68.1&\textbf{69.4}&77.3&77.5&77.5&68.4&66.0&67.2&\textbf{72.2}&70.5& - \\[3pt]
        BERT-base+FocalLoss&59.0&\textbf{76.6}&66.7&78.7&79.9&79.3&60.5&71.5&65.6&66.1&76.0&70.7 \\[3pt]
        RoBERTa-large+augmentation&64.1&68.8&66.3&\textbf{81.0}&83.7&82.3&55.3&74.3&63.4&72.0&76.2&\textbf{74.0}\\[3pt]
        FT+POS+SBS&56.6&71.6&63.2&79.0&\textbf{87.0}&\textbf{82.8}&60.5&74.3&66.7&66.9&\textbf{79.3}&72.6\\[3pt]
        FFA & 56.8&74.2&64.3&\textbf{81.0}&84.5&82.7&55.1&\textbf{77.1}&64.3&67.3&\textbf{79.3}&72.8\\[3pt]
        \bottomrule
          \end{tabular}
  }
\end{table*}
\subsection{Experiment Setup}
We process the utterances and utilize the public pre-trained language models via Transformers \cite{wolf-etal-2020-transformers} from HuggingFace\footnote{\url{https://huggingface.co/models}}. During training, we first initialize the two-stream architecture with Bart\cite{bart} and Funnel Transformer\cite{funnel}, respectively. Adam optimizer is used with default parameters and the learning rate of Bart-large and Funnel Transformer-xlarge are both 5e-6. In terms of self-attention based fusion layer, we adopt 8 attention heads with the hidden size of 3072. Overall, the maximum sequence length is 256, batch size is 8, and the dropout rate in our experiments is set to 0.2, we also introduce R-Drop \cite{DBLP:journals/corr/abs-2106-14448} to act as a regularizer. Moreover, we adopt early stopping with a patience of 8 epochs to avoid overfitting. We train and evaluate all the models on the 32GB Tesla V100.

\subsection{Overall Results}
Table \ref{tab:1} presents the results of FFA compared to top performers. Apparently, the transformer-based approaches are far superior to those RNN-based. Our method accomplishes the best results in the Comma (precision, F1-score), the Question (recall), and the overall (F1-score), separately. In contrast, ELECTRA-large+Disc-ST obtains high recall but low precision compared to ours. FFA exceeds FT+POS+SBS in all other aspects except the F1-score of Questions. Thus, the results validate the robustness of our proposed context-aware feature fusion framework.

When analyzing the results in detail, a portion of the models augments the performance by incorporating some extra information. For instance, Self-attention-word-speech utilizes both lexical and prosody features. FT+POS+SBS employs POS tools to add additional POS tags and bring POS knowledge for punctuation restoration. RoBERTa-large+augmentation enhances data through insertion, substitution, and deletion. Unlike these methods, FFA leverages the training dataset itself and improves 0.9\%, 2.3\%, and 12.3\% compared to FT+POS+SBS, RoBERTa-large+augmentation, and Self-attention-word-speech, respectively. In other words, it can reveal that our approach could further advance via data augmentation or external knowledge.

Compared to the current SOTA model ELECTRA-large+Disc-ST, it's a discriminative self-training approach, which extends the training dataset (2M) with a large amount of unlabeled data (30M), the new dataset is nearly 15 times larger than the training set. It applies a teacher model to generate the pseudo labels and yield the final student model. The key differences lie on whether exploiting external data, we mainly focus on advancing the transformer-based architecture by introducing the interaction and masked self-attention based modules. The experiments illustrate our approach can achieve competitive performance without external data.

In terms of the results on the ASR transcription test set. Without focal loss or data augmentation, our approach outperforms those bert-based models by a large margin. FFA obtains three best results in total, 81.0\% in precision of period, 79.1\% in recall of question, and 79.3\% in precision of overall, respectively.  Compared to the SOTA model RoBERTa-large+augmentation in overall performance, FFA has a slight lower precision and a competitive F1-score, but leads to 3.1 points improvement in recall. One possible reason is that the extra data can benefit the robustness, especially when large quantities of noise samples exist in the transcription set.

\begin{table}[!ht]
  \caption{The results of ablation study.}
  \label{tab:3}
  \centering
  \setlength{\tabcolsep}{2.8mm}{
    \begin{tabular}{ l|r|r|r}
       \toprule
        \multirow{2}*{\textbf{Model}}&\multicolumn{3}{c}{\textbf{Overall}} \\
        \cline{2-4} & P & R& F1\\
        \hline
        FFA&\textbf{84.2}&\textbf{86.1}&\textbf{85.2}\\[2pt]
           - Interaction &	83.8&	85.8&	84.8\\[2pt]
           - Interaction \& fusion & 83.2&	85.1&	84.2\\[2pt]
           - Interaction \& fusion \& R-Drop &	83.2&	84.9&	84.0\\[2pt]
        \bottomrule
          \end{tabular}
  }
\end{table}

\subsection{Ablation study}
We perform a thorough ablation study to show the contribution of each design choice.

First, the results in the bottom of table \ref{tab:1} demonstrate that the performance has a large drop while eliminating ISA (3\%) or MSA (1.7\%), respectively. It verifies the two-stream architecture is more effective and robust. Second, we evaluate FFA by replacing the interaction attention with multi-head attention, the precision, recall and F1-score decrease 0.4, 0.3 and 0.4 points, respectively (shown in table \ref{tab:3}). Then, we directly remove the fusion layer and concatenate the two embeddings. The performances have a slight drop from 83.8\% to 83.2\%, 85.8\% to85.1\%, and 84.8\% to 84.2\%. The metrics on F1-score decreases 0.2 points when we drop out R-Drop. The studies confirm that the design of FFA can boost the effectiveness.

\section{Conclusions}
In this paper, we propose a novel context-aware feature fusion framework (FFA) based on two-type self-attention mechanism. By the interaction and masked self-attention based modules, our framework not only pays more attention to the feature representations of utterances, but has the context-aware ability. The experiments on the benchmark dataset IWSLT demonstrate that our method is more effective than previous works. Since we do not exploit the external knowledge or data augmentation, future work can leverage extra information to further boost the inference performance.

\bibliographystyle{IEEEtran}

\bibliography{mybib}
\end{document}